\documentclass[lettersize,journal]{IEEEtran}
\usepackage{amsmath,amsfonts}
\usepackage{algorithmic}
\usepackage{algorithm}
\pdfoutput=1
\usepackage{array}
\usepackage[caption=false,font=normalsize,labelfont=sf,textfont=sf]{subfig}
\usepackage{textcomp}
\usepackage{stfloats}
\usepackage{url}
\usepackage{verbatim}
\usepackage{graphicx}
\usepackage{cite}
\usepackage{color}
\usepackage{float}
\usepackage{multirow}
\usepackage{stfloats}
\usepackage{booktabs}
\usepackage{threeparttable}
\begin{document}

\title{Improving Critical Node Detection Using Neural Network-based Initialization in a Genetic Algorithm}

\author{Chanjuan Liu,~\IEEEmembership{Member,~IEEE,} Shike Ge, Zhihan Chen, Wenbin Pei, Enqiang Zhu$^*$,\\
Yi Mei,~\IEEEmembership{Senior Member,~IEEE,} Hisao  Ishibuchi,~\IEEEmembership{Fellow,~IEEE}
\thanks{This work was supported in part by the National Natural Science Foundation of China under Grants (62172072,62272115), in part by the Natural Science Foundation of Guangdong Province of China under Grant 2021A1515011940, and in part by Science and Technology Projects in Guangzhou. (Corresponding author: Enqiang Zhu.) }
\thanks{C. Liu, S. Ge and P. Wen are with the School of Computer Science and Technology, Dalian University of Technology, 116024, China.}
\thanks{Z. Chen is with the Institute of Software, Chinese Academy of Science, Beijing 100190, China.}
\thanks{E. Zhu is with the Institute of Computing Science and Technology, Guangzhou University, Guangzhou 510006, China.}
\thanks{Y. Mei is with the School of Engineering and Computer Science, Victoria University of Wellington, Wellington 6012, New Zealand}
\thanks{H. Ishibuchi is with the Guangdong Provincial Key Laboratory of Brain-Inspired Intelligent Computation, Department of Computer Science and Engineering, Southern University of Science and Technology, Shenzhen 518055, China.}
}

\markboth{Journal of \LaTeX\ Class Files,~Vol.~14, No.~8, August~2021}%
{Shell \MakeLowercase{\textit{et al.}}: A Sample Article Using IEEEtran.cls for IEEE Journals}


\maketitle

\begin{abstract}
The Critical Node Problem (CNP) is concerned with identifying the critical nodes in a complex network. These nodes play a significant role in maintaining the connectivity of the network, and removing them can negatively impact network performance. CNP has been studied extensively due to its numerous real-world applications. Among the different versions of CNP, CNP-1a has gained the most popularity. The primary objective of CNP-1a is to minimize the pair-wise connectivity in the remaining network after deleting a limited number of nodes from a network. Due to the NP-hard nature of CNP-1a, many heuristic/metaheuristic algorithms have been proposed to solve this problem. However, most existing algorithms start with a random initialization, leading to a high cost of obtaining an optimal solution. To improve the efficiency of solving CNP-1a, a knowledge-guided genetic algorithm named K2GA has been proposed.  Unlike the standard genetic algorithm framework, K2GA has two main components: a pretrained neural network to obtain prior knowledge on possible critical nodes, and a hybrid genetic algorithm with local search for finding an optimal set of critical nodes based on the knowledge given by the trained neural network.  The local search process utilizes a cut node-based greedy strategy. The effectiveness of the proposed knowledge-guided genetic algorithm is verified by experiments on 26 real-world instances of complex networks. Experimental results show that K2GA outperforms the state-of-the-art algorithms regarding the best, median, and average objective values, and improves the best upper bounds on the best objective values for eight real-world instances.

\end{abstract}


\section{Introduction}
\IEEEPARstart{W}{ith} the rapid development of information science and technology in the last two decades, artificial systems have become more complex, having many dynamic and nonlinear characteristics caused by complex internal interactions and multidimensional attributes of system components \cite{1998Dynamics}. Typical complex systems include the Internet of Things \cite{2012Interconnectedness}, disease spread networks \cite{2014A}, multi-agent systems \cite{WangW22}, and complex networks \cite{2006Identifying,00030HLS21}. 

\subsection{Background}

It has been common to model a complex system into a graph where nodes represent system components, and edges represent the internal interaction between nodes. In a complex system, some of the components can play a critical role in the maintenance of system performance, and these components are called critical nodes. For instance, if the performance of a system deteriorates when a node fails or when it is attacked, then that node is considered a critical node. Thus, by protecting or controlling the critical nodes of a system,  the internal interactions and dynamic behavior of the system can be manipulated at a minimum cost.
Critical nodes have been introduced by different names in related studies, such as most vital nodes \cite{1982Most}, key-player nodes \cite{2006Identifying},  most influential nodes \cite{ZHU2024127195,9434427}, and most $K$-mediator nodes \cite{2011Finding}.
This paper focuses on the network connectivity problem, where the term ``critical nodes'' is used. 

Given a graph $G=(V, E)$, a critical node problem (CNP) aims to find a subset $S\subseteq V$, called \emph{the critical node set}, such that $G-S$ has the worst connectivity, where  $G-S$ is obtained from $G$ by deleting nodes in $S$ and their incident edges.

\subsection{Significance}

The critical node problem is significant both in theory and in practice \cite{LIU2023119140}.  In theory, the CNP is shown to be an $\mathcal{NP}$-hard problem, which is of great importance for theoretical study. In practice, the CNP has extensive applications in complex network analysis, such as drug design \cite{jhoti2007structure,stromgaard2009textbook}, disease prediction \cite{tomaino2012studying}, epidemic spread control \cite{tao2006epidemic}, and key player identification \cite{2006Identifying}.
Here, the largest social network platform, Facebook, is used as an example. The network is modeled as a graph $G=(V, E)$, where a node set $V$ represents a set of Facebook users, and edges in the edge set $E$ indicate the interactions between users, such as browsing, thumbs-up, commenting, and following each other. Critical nodes in $G$ (i.e., critical users in the network)  have a pivotal role in the connectivity maintenance of $G$, and their influence can spread over the whole network. Thus,
business cooperation, advertising, and other commercial marketing activities should be conducted against critical users to maximize the benefits. Given the significance of CNP, it is crucial to design efficient algorithms for solving the CNP.

\subsection{Challenge}

It should be noted that different connectivity metrics imply different CNPs. There are three main types of connectivity metrics involved in CNPs: (a) the number of connected pairs of nodes in $G-S$; (b) the number of nodes in the maximum connected component of $G-S$; (c) the number of connected components of $G-S$. We use $\sigma_a$, $\sigma_b$, and $\sigma_c$ to denote these three connectivity metrics, respectively.  In addition, according to different constrained formulations, CNPs can be roughly divided into two classes denoted by CNP-1 and CNP-2. Class CNP-1 aims to optimize (maximize or minimize) a predefined network connectivity metric under a constraint condition on the number of deleted nodes. Class CNP-2 aims to minimize the number of deleted nodes under a constraint condition on the value of a connectivity metric. Different combinations of classes \{CNP-1, CNP-2\} and metrics \{$\sigma_a$, $\sigma_b$, $\sigma_c$\} provide the classification of CNPs into six problems as shown in Table \ref{tab:table1}. This paper aims to explore a method for the most studied CNP-1a problem.

\begin{table}[H]
\renewcommand\tabcolsep{18pt}
\renewcommand{\arraystretch}{1.2}
\caption{Classification for CNPs\label{tab:table1}}
\centering
\begin{tabular}{|c|c|c|}
\hline
Connectivity metrics & CNP-1 & CNP-2\\
\hline
$\sigma_a$  & CNP-1a & CNP-2a\\
\hline
$\sigma_b$  & CNP-1b & CNP-2b\\
\hline
$\sigma_c$  & CNP-1c & CNP-2c\\
\hline
\end{tabular}
\end{table}

Considering the $\mathcal{NP}$-hard characteristic of CNP-1a, the exact algorithms for solving  CNP-1a require a long computation time and are mainly suitable for small networks or networks with special structures.
Accordingly, many heuristic approaches have been proposed for solving CNP-1a. However, the majority of the existing heuristic algorithms for solving CNP-1a use random initial solutions, which leads to a lower quality guarantee of the initial solutions and thus affects the quality of the final solutions.

\subsection{Main contribution}

To address the issue of random initialization in the standard framework of genetic algorithms for CNP-1a, in this study, we propose a knowledge-guided genetic algorithm, called K2GA. Unlike the standard GA framework \cite{687888}, K2GA is initialized by a pretrained graph neural network (GNN) model. The GNN model is designed for processing graph information, where a real-world graph is embedded into a low-dimensional space according to the structure information of the graph, node features, and task requirements \cite{zhou2022graph}. 


Inspired by the effectiveness of GNNs in graph-related learning tasks (e.g., node classification, graph classification, and link prediction and clustering), the proposed GNN model is trained to obtain prior knowledge on potential critical nodes. Taking the prediction of critical nodes as initial solutions, a local search-based genetic algorithm is then used to find an optimal solution. 
Equipped with a local search process and population-based operations, the proposed genetic algorithm searches for an optimal set of critical nodes based on the guidance of the GNN. A cut node-based greedy strategy is utilized as heuristic information in the local search, which minimizes the number of the connected node pairs to facilitate the movement to the optimal objective value.

The proposed K2GA algorithm is evaluated on  26 real-world benchmark instances of complex networks. Compared with state-of-the-art heuristic algorithms for solving CNP-1a, the proposed K2GA can achieve optimal results for most instances and has the best performance with respect to the best objective value, median objective value, and average objective value.
Moreover, K2GA is compared with its two variants, K2GA\_randInit and K2GA\_randInit\_longImprove. K2GA\_randInit is the same as K2GA except for a random initialization of the population; K2GA\_randInit\_longImprove also adopts random initialization, but it uses a heuristic algorithm to improve the initial solutions in the initial population with a longer time limit (i.e., 24 hours).
Experimental results indicate that K2GA outperforms the two variants in solution quality. These results further prove the effectiveness of the framework of guiding genetic algorithms via prior knowledge in solving CNP-1a.

\subsection{Paper structure}
The remainder of this paper is organized as follows. Section \ref{sec-2} presents a brief review of related work. Section \ref{set-pre}  provides preliminary information, after which we describe the proposed K2GA algorithm in Section \ref{sec-3}. Section \ref{sec-4} reports experimental results. Finally, Section \ref{sec-5} concludes the paper with future research.

\section{Related Work}\label{sec-2}

\subsection{Evolutionary and Local Search Algorithms for CNPs}

The greedy strategy and the local search framework are extensively utilized for solving CNP-1a. One of the typical algorithms of this category is the Greedy1 algorithm, which was proposed by Arulselvan et al. \cite{arulselvan2009detecting} and is the earliest greedy-based algorithm for CNP-1a. It constructs the complement of a critical node set by extending an initial maximal independent set. Later, Ventresca and Aleman \cite{ventresca2014fast} proposed another greedy-based algorithm, named Greedy2, for solving CNP-1a, which greedily removes nodes until a feasible solution is obtained. In 2015, Pullan \cite{pullan2015heuristic} introduced the idea of greedily deleting and selecting nodes from large connected components to solve CNP-1a. To improve the efficiency of Pullan's algorithm,  Addis et al.\cite{addis2016hybrid} proposed two CNP-1a algorithms, named Greedy3d and Greedy4d, which search the solution space by alternately deleting and adding nodes to the current solution.  In 2016, Aringhieri et al. \cite{aringhieri2016local}  proposed a local search engine and combined it with the iterative local search (ILS) in \cite{Lourenço2003} and the variable neighborhood search (VNS) in \cite{Hansen} to solve CNP-1a.  Although the local search-based algorithm can obtain a solution in a short time, its ability to solve CNP-1a  on large graphs has been limited by the local optimum problem.


Considering the advantages of evolutionary algorithms (EAs) in solving combinatorial optimization problems, Aringhieri et al. \cite{ARINGHIERI2016359} proposed a  genetic algorithm to solve CNP-1a and was shown to outperform the heuristic method. In the same year, Aringhieri et al.\cite{ARINGHIERI2016128} proposed a general genetic framework to solve various classes of CNPs \cite{ARINGHIERI2016128}. To enhance the optimization process of EAs, Zhou et al. \cite{zhou2018memetic} proposed a memetic algorithm (MA), named MACNP. Memetic algorithms typically combine local search heuristic algorithms with EAs and have been proven to be more effective than traditional EA algorithms on CNP-1a \cite{krasnogor2005tutorial}. Recently, Zhou et al. \cite{zhou2020variable} introduced a variable population mechanism to the MA framework and proposed a variable population memetic search (VPMS) method for CNP-1a. VPMS performs the best among the existing heuristic algorithms for solving CNP-1a. However, these state-of-the-art algorithms generate the initial population randomly. Once an initial assignment falls into an unpromising search space, the evolutionary algorithm has difficulty in obtaining an optimal solution. Thus, an important research issue is to design an effective initialization mechanism for boosting CNP-1a solving.

\section{Preliminaries} \label{set-pre}

This section presents preliminary information related to this study, including the problem description and the graph attention (GAT) network.

\subsection{Notations}
All graphs considered in this study are finite, undirected, and simple. Given a graph $G=(V, E)$ with the \emph{node set} $V$ and the \emph{edge set} $E$, deleting a set $S\subseteq V$ of nodes removes all nodes in $S$ and their incident edges from $G$, which represents the resulting graph denoted by $G-S$. For a graph $G$, $V(G)$ and $E(G)$ denote the node and edge sets of $G$, respectively. Two nodes $u$ and $v$ are considered \emph{adjacent} in $G$ if $uv\in E(G)$, and two adjacent nodes are considered \emph{neighbors}. We denote a set of neighbors of a node $u$ in $G$ by $N_G(u)$ (or simply by $N(u)$ when $G$ is fixed),  i.e., $N_G(u)= \{v|uv\in E(G)\}$. The \emph{degree} of a node $u$ in $G$, which is denoted by $d_G(u)$, represents the number of neighbors of $u$ in $G$, i.e., $d_G(u)=|N_G(u)|$. A \emph{subgraph} of $G$ is a graph $H$ that satisfies the conditions of $V(H) \subseteq V(G)$ and $E(H)\subseteq E(G)$. Two nodes are considered \emph{connected} in a graph if there is a sequence of alternating nodes and edges that connect them. 
A \emph{connected component} of $G$ is a subgraph $G'$ of $G$ such that any pair of nodes are connected in $G'$ and $N_{G}(u) \subseteq V(G')$ for all $u\in V(G')$. A graph is considered \emph{connected} if it contains only one connected component. The \emph{distance} between nodes $x$ and $y$ in a connected graph $G$, denoted by $dis_G(x,y)$, represents the number of edges along the shortest path connecting the nodes $x$ and $y$.

There are several node features characterizing the importance of nodes. 

$\bullet$ The \emph{closeness centrality}  \cite{bavelas1950communication} measures the closeness of a node to the other nodes in a network, and it describes the broadcaster role of the node in a network. In a connected graph $G$,  the  \emph{closeness centrality} $C(v)$ of a node $v\in V(G)$ is given by

\begin{equation}\label{closeness}
C(v)=\frac{|V(G)|-1}{\sum\limits_{u\in V(G)} dis_G(v,u)}.
\end{equation}

$\bullet$ The \emph{betweenness centrality} measures an individual's control over communication between others in a social network. It was introduced by Freeman in 1977. In a connected graph $G$, the betweenness centrality $C_B(v)$ of a node $v\in V(G)$ is given by
\begin{equation}\label{betweenness}
C_B(v)= \sum\limits_{u\neq v \neq w \in V(G)} \frac{\sigma_{uw}(v)}{\sigma_{uw}},
\end{equation}
where $u$, $w$ are a fixed pair of nodes in $V(G)$, $\sigma_{uw}$ is the number of shortest paths between $u$ and  $w$, and $\sigma_{uw}(v)$ is the number of shortest paths between $u$ and  $w$ that pass through $v$.

$\bullet$ The \emph{degree centrality} $C_D(v)$ of a node $v\in V(G)$ represents  the degree of $v$, and it is calculated as follows:
\begin{equation}\label{degree}
C_D(v)= d_G(v).
\end{equation}

$\bullet$ The \emph{clustering coefficient} was introduced by Watts and Strogatz \cite{1998Collective}, and it is used to quantify how close the neighbors of a node are to a complete graph.  For a  graph $G$ and a node $v\in V(G)$, assume that $G'= G- (V(G)\setminus N_G(v))$; then, the clustering coefficient $C_C(v)$ of a node $v\in V(G)$ is calculated as follows:

\begin{equation}\label{clustering}
C_C(v)= \frac{2|E(G')|}{|V(G')|(|V(G')|-1)}.
\end{equation}\
\subsection{Problem Description}

Given a graph G=(V, E) and a subset of nodes $S \subseteq V(G)$, the residual graph after deleting $S$ can be represented by a set of disjoint connected components $H=\{C_1, C_2, \dots, C_L\}$, where each $C_i$ ($1\geq i \geq L$) is a $i$-th connected component. The objective of CNP-1a is to remove a subset of nodes $S$ with $|S| \leq k$ ($k$ is a given integer), such that the pairwise connectivity of the residual graph is minimized, i.e., the number of connected node pairs in $H$ is minimized. Formally, the objective function $f(S)$ of CNP-1a is defined as:
\begin{equation}\label{equ-1}
f(S)=\sum_{i=1}^{L}\frac{|C_i|(|C_i|-1)}{2},
\end{equation}
where $L$ represents the number of connected components in $H$, and $|C_i|$ denotes the number of nodes in the $i$-th connected component $C_i$.

The input and output of CNP-1a are formally described as:

\textbf{Problem (CNP-1a):}

\textbf{Input:} An undirected graph $G =(V,E)$ and an integer $k$.

\textbf{Output:} A solution $S$ with $|S| \leq k$, which minimizes $f(S)$.

\subsection{GAT Model}
Convolutional neural networks (CNNs) have been effectively used in the image and text domains \cite{LIN2024202}, with a 'kernel' to extract certain 'features' from an input image in Euclidean space. However, there are many irregular data structures in the real world such as traffic networks, power networks, and protein networks. For these structures, the graph convolutional network (GCN) \cite{kipf2016semi,LiuSCIS2023} applies convolutional operations to graph data, i.e., it learns the representation of node $v_i$ by aggregating its features $h_i$ with its neighbors' features $h_j$. However, the neighbors of a node do not have the same importance and may also contain invalid information. To address these issues, the graph attention (GAT) network \cite{velickovic2017graph} introduces an attention mechanism that assigns different attention scores and weights to neighbors. Moreover, the GAT does not rely on structural information of the whole graph, which makes the GAT suitable for inductive tasks. The GAT network is constructed by stacking GAT layers, where each GAT layer maps an $F$-dimensional feature set $\mathbf{h}=\left\{\vec{h}_{1}, \vec{h}_{2}, \ldots, \vec{h}_{N}\right\}, \vec{h}_{i} \in \mathbb{R}^{F}$ to an $F^\prime$-dimensional feature set $\mathbf{h}^{\prime}=\left\{\vec{h}_{1}^{\prime}, \vec{h}_{2}^{\prime}, \ldots, \vec{h}_{N}^{\prime}\right\}, \vec{h}_{i}^{\prime} \in \mathbb{R}^{F^{\prime}}$; That is, the dimensionality of the feature vector is changed from $F$ to $F^\prime$ by the data processing in the GAT. The attention coefficient $a_{ij}$ needs to be calculated for each node pair $(i,j)$ in GAT layers as follows:
\begin{equation}\label{equ-2}
\alpha_{i j}=\frac{\exp \left(\operatorname{LeakyReLU}\left(\overrightarrow{\mathbf{a}}^{T}\left[\mathbf{W} \vec{h}_{i} \| \mathbf{W} \vec{h}_{j}\right]\right)\right)}{\sum_{k \in N(i)} \exp \left(\operatorname{LeakyReLU}\left(\overrightarrow{\mathbf{a}}^{T}\left[\mathbf{W} \vec{h}_{i} \| \mathbf{W} \vec{h}_{k}\right]\right)\right)},
\end{equation}
where $W \in R^{F^{\prime} \times F}$ is the linear transformation weight matrix,   $\overrightarrow{\mathrm{\mathbf{a}}} \in \mathbb{R}^{2 F^{\prime}}$ represents the feedforward neural network's parameters, and LeakyReLU with a negative semiaxis slope of 0.2 is used as a nonlinear activation function.  The feature set of a node $i$ is computed as follows:

\begin{equation}\label{equ-3}
\vec{h}_{i}^{\prime}=\sigma\left(\sum_{j \in N(i)} \alpha_{i j} \mathbf{W} \vec{h}_{j}\right),
\end{equation}
where $\sigma$ is a nonlinear activation function, and $j\in N(i)$ indicates traversing all the neighbors $j$ of $i$. In addition, the GAT uses a multiheaded attention mechanism  \cite{vaswani2017attention} to keep the learning process stable.

\section{Knowledge-guided Genetic Algorithm}\label{sec-3}
This section introduces the proposed knowledge-guided genetic algorithm (K2GA) for solving CNP-1a. When using genetic algorithms to solve CNP-1a, a traditional approach to construct the initial population is to select some nodes from the graph randomly. However, the initial solutions (individuals) generated in a completely random way may have poor quality. Generally, a high-quality population can guide a genetic algorithm to obtain a better solution and improve the convergence speed of the algorithm.

Therefore, this study combines a pretrained GNN with a genetic algorithm that uses the trained GNN to generate a set of promising candidate nodes. Initial solutions for the genetic algorithms are generated by randomly selecting nodes from the promising candidate node set. In this manner, we create a high-quality initial population, which is much better than a pure random initial population. The generated initial population efficiently guides the search of the proposed local search-based genetic algorithm towards the optimal solution. More specifically, crossover generates other subsets of the predicted promising candidate node set, and local search tries to improve each solution in the current population. It should be noted that the inclusion of non-predicted nodes is examined in local search in addition to the inclusion of predicted nodes. In this manner, we prevent the local search-based genetic algorithm from focusing only on the predicted nodes too much.


\subsection{Proposed Algorithm Framework} \label{sec-3.A}

\begin{figure}[!t]
\centering
\includegraphics[width=7cm]{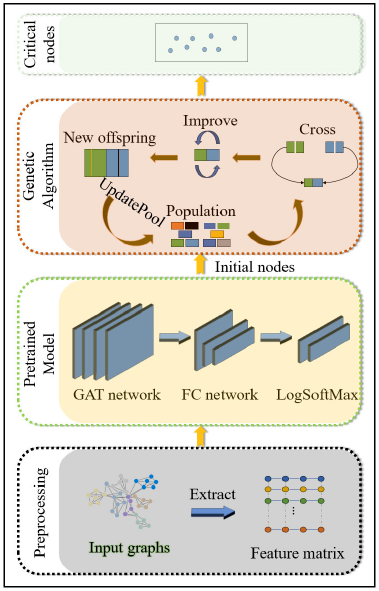}
\caption{Flowchart of K2GA}
\label{fig_1}
\end{figure}

The flowchart and pseudocode of the proposed K2GA algorithm are shown in Figure \ref{fig_1} and Algorithm \ref{alg:alg1}, respectively. 
 
In Figure \ref{fig_1}, a graph and the feature matrix of this graph are taken as the input to the pretrained GNN model. The model outputs a set of predicted initial nodes, based on which an initial population is formed for the genetic algorithm. The genetic algorithm searches for the optimal solution iteratively. Each iteration consists of four processes: (1) selection; (2) crossover; (3) improvement by local search; and (4) population updating. The algorithm terminates when the time limit is reached.

As shown in Algorithm \ref{alg:alg1},  a set of potential critical nodes is first predicted by a trained neural network \textsc{Model}, which accepts a graph $G$ and its feature matrix $FeaMax$ as input
(line 3). Based on the predicted nodes,  a procedure \textsc{InitialPop} (see Section \ref{aa2}) is used to generate an initial population $\textbf{P}$ (line 4). Assume that $S^\ast$ is the current best solution in $\textbf{P}$ (line 5). Then, the algorithm performs an iteration (the main genetic processes) as follows. It randomly selects two solutions $S_i$ and $S_j$ from the current population set $\textbf{P}$ and uses a crossover operator \textsc{Cross} (see Section \ref{aa3}) to generate an offspring solution $S$ based on the parent solutions $S_i$ and $S_j$ (lines 7 and 8). Next, a local search procedure \textsc{Improve} (see Section \ref{aa4}) is conducted to improve $S$, and when a better solution $S'$ than $S^{\star}$ is found, the current best solution $S^\ast$ is replaced by $S'$, and a procedure \textsc{UpdatePop} (see Section \ref{aa5}) is performed to update the current population $\textbf{P}$ (lines 9-13). We do not use any mutation operator since the local search procedure is used to examine mutated solutions. Finally, when the loop reaches its time limit, the algorithm returns the current optimal solution $S^{\star}$ (line 15).

According to line 3 Algorithm \ref{alg:alg1}, K2GA uses a pretrained neural network to generate initial solutions. By replacing this line with  ``$Ini\_nodes$ $\leftarrow$ $k$ nodes randomly selected from $V(G)$", we obtain a variant of K2GA called the K2GA\_randInit algorithm. That is, K2GA\_randInit randomly selects $k$ nodes as an initial solution. K2GA\_randInit is used to generate a good solution for each instance in the datasets for training the neural network used in K2GA, which is performed as an off-line procedure (see Section IV-C for more details).


\begin{algorithm}[!h]
\caption{Pseudo-Code of the K2GA Algorithm}\label{alg:alg1}
\begin{algorithmic}[1]
\STATE \textbf{Input:} An undirected graph $G = (V,E)$ and an integer $k$
\STATE \textbf{Output:} The best solution $S^{\star}$ found so far
\STATE $Ini\_nodes$ $\gets$  {\textsc{model}}$(G, FeaMax)$
\STATE $\textbf{P}$ $\gets$ {\textsc{InitialPop}} ($Ini\_nodes$, $k$)
\STATE $S^{\star} \gets \textbf{argmin}_{S \in {\textbf{P}}}f(S)$
\WHILE{$runningtime< cutoff$}
\STATE Randomly select two solutions $S_i, S_j$ from \textbf{P};
\STATE $S \gets \textsc{Cross}(S_i, S_j, k)$;
\STATE $S' \gets \textsc{improve} (S)$;
\IF{$f(S')<f(S^{\star})$}
\STATE $S^{\star}  \gets S'$;
\STATE $\textsc{UpdatePop}(\textbf{P}, S')$;
\ENDIF
\ENDWHILE
\STATE \textbf{return}  $S^{\star}$
\end{algorithmic}
\end{algorithm}

\subsection{Neural Network Model} \label{aa1}
A pretrained GNN-based approach {\textsc{model}} is proposed to predict potential critical nodes. As shown in Figure \ref{fig_1}, the core of {\textsc{model}} is a neural network model that consists of a four-layer GAT network and a fully connected (FC) network. The input of {\textsc{model}} includes a graph $G$ and a feature matrix of $G$, and the output consists of possible critical nodes.

\subsubsection{Input}

The neural network model takes a graph and its corresponding feature matrix as input. When calculating the feature matrix, to make it easier to calculate the weighted index on different orders of magnitude, we follow the normalization method in \cite{ZHAO202018}. Specifically, feature $\alpha$ is normalized as:

\begin{equation}\label{normalize}
f_{\alpha}(v) = \frac{r_{\alpha}(v)}{n} -0.5,
\end{equation}
where $n$ is the number of nodes in the graph and $r_{\alpha}(v)$ denotes the ranking position of a node $v$ in the network according to
the value of feature $\alpha$.

\subsubsection{GAT Network and Fully Connected Network}

The model employs a Graph Attention Network (GAT) to acquire knowledge about node features. In contrast to the Graph Convolutional Network (GCN) model that extracts information from the entire graph structure, the GAT requires only feature data of a node and its neighboring nodes. Hence, GAT is more appropriate for inductive learning \cite{velickovic2017graph}. To accomplish the inductive learning task, a four-layer GAT network is utilized.

The GAT network has hidden layers with a filter size of $64\times |V(G)|$. The number of hidden layers and neurons in them is determined by the random search algorithm described in Section \ref{bb1}. The ELU activation function \cite{clevert2015fast} is used in these hidden layers, which is defined as follows:

\begin{equation}
ELU(x) = \max(0,x)+\min(0, \lambda(e^x-1)),
\end{equation}
where $\lambda$ is a parameter, which is set to 1 in general.

The model employs a three-layer fully connected network to classify the features learned from the GAT network and to classify network nodes. The fully connected network uses the ELU activation function. The LogSoftMax classifier is used to process the outputs of the fully connected network, which is defined as follows:

\begin{equation}
\operatorname{LogSoftmax}\left(x_{i}\right)=\log \left(\frac{\exp \left(x_{i}\right)}{\sum_{j=1}^{n} \exp \left(x_{j}\right)}\right),
\end{equation} 
the exponential function is applied to each element $x_i$ of the input vector $\mathbf{x}$, where $\mathbf{x} = (x_1, . . . , x_n)$.

\subsection{Model Pretraining}\label{bb1}
To improve the accuracy of a neural network model for predicting high-quality critical nodes, it is necessary to obtain a set of appropriate values of network hyperparameters \cite{9359655}.  Nevertheless, manually tuning hyperparameters can be a complex and time-consuming process. Therefore, this study uses random search to tune hyperparameters automatically. Even though  random search may not optimize hyperparameters, it is simple and effective. Moreover, it is acceptable that a neural network model cannot guarantee optimality. This is because the role of the neural network model here is only to select important nodes to generate a good initial population for the genetic algorithm, as shown in Figure \ref{fig_1}.  We spent approximately 24 hours on hyperparameter tuning in the neural network using the random search method.

\begin{figure*}[!t]
\centering
\includegraphics[width=15cm]{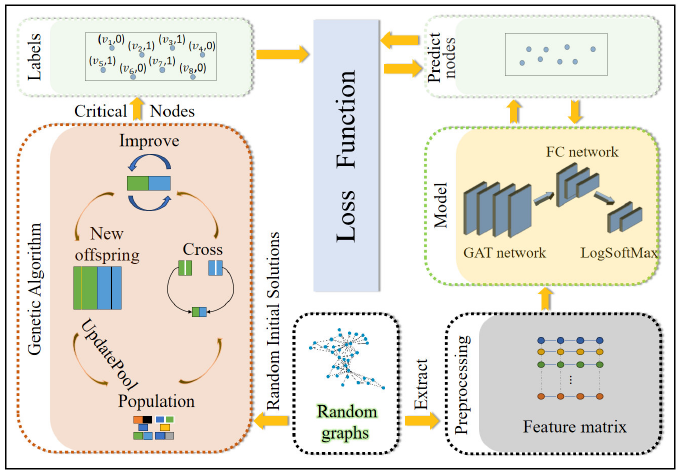}
\caption{Training process of K2GA}
\label{train}
\end{figure*}

For the pre-training of a neural network model that works with various graphs, 300 graphs are randomly generated (for details about how to generate those graphs in our computational experiments).The process of pre-training the neural network is depicted in Figure \ref{train}. Note that the genetic algorithm shown in Figure \ref{train} is distinct from the one in Figure \ref{fig_1}. The genetic algorithm in Figure \ref{fig_1} is used to search for optimal solutions for solving CNP instances, whereas the one in Figure \ref{train} is the genetic algorithm with a random initial population, called K2GA\_randInit (as described in Section \ref{sec-3.A}), which is utilized to find a set of critical nodes for each of the 300 graphs.

Since K2GA\_randInit is a stochastic search method and depends on the initial population, different solutions can be obtained from different runs on the same graph. Therefore, 10 runs of the genetic algorithm were performed on each graph, resulting in 10 solutions for each graph. To obtain a target node set, which can be seen as a set of promising nodes, the union of all nodes in the 10 solutions for each graph was used. Thus, an input-output pair for the pre-training of the neural network was generated by pairing each graph with its corresponding promising node set. In this way, a total of 300 input-output pairs were created for the pre-training of the neural network.


The $k$ value in the CNP-1a problem considered in this study is defined as follows:
\begin{equation}\label{equ-k-value}
k= \frac{n}{n_C}\times 0.3,
\end{equation}
where $n$ is the number of nodes in a graph, and $n_c$ is the number of connected components in the graph. In each training iteration,  the critical nodes obtained by the neural network model are compared with the labels obtained by K2GA\_randInit, and the neural network is trained using the negative log-likelihood loss function and Adam optimizer \cite{2014Adam}. Finally, the trained model can be used to
predict critical nodes in a new graph (i.e., to create a good initial population for the genetic algorithm in Figure 1).

\subsection{The Cut Node-Based Greedy Strategy}

 For a node $v\in V$, let $G-v$ be the resulting graph obtained from $G$ by removing $v$ and its incident edges. If $G-v$ has more connected components than $G$  (i.e., if the removal of $v$ increases the number of connected components), then $v$ is called a cut node of $G$. 

{\bf Theorem 1.} Let $u$ and $v$ be two nodes in a connected graph $G$.  Suppose that  $G-u$ and $G-v$ have $p$ and $q$  components, respectively. Denote by   $n_i$ ($i=1,2,\ldots, p$)   and  $m_i$ ($i=1,2,\ldots, q$) the cardinalities of  components of  $G-u$ and  $G-v$, respectively, where   $n_i\geq n_{i+1}$ for $i=1,2,\ldots, p-1$ and $m_i\geq m_{i+1}$ for $i=1,2,\ldots, q-1$.   If $p\leq q$ and $n_i\geq m_i$ for $i=1,2,\ldots, p$, then $t(G,u)\geq t(G,v)$. 

{\emph{Proof}. } 
 If $p=q$, then   $n_i= m_i$ for $i\in \{1,2,\ldots, p\}$ and $t(G,u)= t(G,v)$. In what follows, we assume that $p<q$, and there exists an $i\in \{1,2,\ldots, p\}$ such that $n_i>m_i$. Let $i_{j}, j=1,2,\ldots, \ell$ ($\ell\leq p$) be the subscripts such that  $n_{i_j}>m_{i_j}$, and let $n_{i_j}-m_{i_j}=q_j$. Clearly, $ q_1+q_2+\ldots+q_{\ell} =m_{p+1}+m_{p+2}+\ldots + m_{q} $.
Note that $t(G,u)=\sum_{i=1}^{p} {n_i \choose 2}$ and $t(G,v)=\sum_{i=1}^{q} {m_i \choose 2}$. Therefore,

\vspace{0.1cm}

\hspace{0.3cm} $t(G,u)-t(G,v)$

$=\sum\limits_{j=1}^{\ell} [{n_{i_j} \choose 2} -  {m_{i_j} \choose 2}]$  $-$ $\sum\limits_{i=p+1}^{q} {m_i \choose 2}$ 


$= \frac{1}{2} [\sum\limits_{j=1}^{\ell} (n_{i_j})^2  -  \sum\limits_{j=1}^{\ell} (m_{i_j})^2-   \sum\limits_{i=p+1}^{q} (m_{i})^2]$

$= \frac{1}{2} [\sum\limits_{j=1}^{\ell} q_j   (n_{i_j}+m_{i_j})  -   \sum\limits_{i=p+1}^{q} (m_{i})^2]$

 $\geq  \frac{1}{2} [\sum\limits_{j=1}^{\ell} q_j  (n_{i_j}+m_{i_j})  -  m_{p+1}  (\sum\limits_{i=p+1}^{q}   m_{i}  )]$ 

 $\geq  \frac{1}{2} [2m_{p+1} (\sum\limits_{j=1}^{\ell} q_j) -  m_{p+1}  (\sum\limits_{i=p+1}^{q}   m_{i}  )] $

$>0$.   \hspace{7cm}$\square$

For any two nodes $u$ and $v$ such that $u$ is a cut node and $v$ is not a cut node, $G-v$ has only one component and $G-u$ has at least two components, and no component in $G-u$ is larger than  $G-v$. Therefore,  if $G$ contains a cut node, then by Theorem 1, the node $x$ that minimizes $t(G,x)$ must be a cut node.  In addition, Theorem 1 provides guidance for identifying a better cut node when $G$ contains multiple cut nodes.

\subsection{Genetic Algorithm}

The genetic algorithm for solving the CNP-1a problem includes four processes: population initialization, crossover, population updating, and local search, where the local search is the core of the algorithm, i.e., the local search is responsible for improving the quality of a single solution (an individual in the population). The genetic algorithm searches for good solutions in the current population in each iteration and outputs the best solution when the algorithm stops.

\subsubsection{Population Initialization} \label{aa2}
Given the pretrained neural network model, the initial population $\textbf{P}$ can be designed based on the set of nodes $Ini\_nodes$ generated by the model, and the pseudocode of this procedure \textsc{InitialPop} is shown in Algorithm \ref{alg:initial}. This procedure generates each individual of $\textbf{P}$ by randomly selecting $k$ nodes from $Ini\_nodes$ if $|Ini\_nodes|>k$; otherwise, it randomly selects $k-|Ini\_nodes|$ nodes from $V(G)\setminus Ini\_nodes$ and adds them to $Ini\_nodes$ to form a solution. {Then we use \textsc{improve} described in Algorithm \ref{alg:impr} to improve the solution.}  Note that when $|Ini\_nodes|=k$, the initial solution is equivalent to $Ini\_nodes$, but diversity will later be introduced by the \textsc{improve} process. 

\begin{algorithm}[H]
\caption{$\textsc{InitialPop()}$}\label{alg:initial}
\begin{algorithmic}[1]
\STATE \textbf{Input:} A set $Ini\_nodes$ of nodes and an integer $k$
\STATE \textbf{Output:} Initial population $\textbf{P}$
\WHILE{$|\textbf{P}| < pop\_size$}
\STATE $S' \gets \emptyset$;
\IF{ $|Ini\_nodes| > k$}
\STATE $S' \gets $ \{Randomly select $k$ nodes from $Ini\_nodes$\};
\ELSE
\STATE $S' \gets  Ini\_nodes$;
\WHILE{$|S'| < k$}
\STATE $S'\cup $ \{a random node from $V(G)\setminus S'$\};
\ENDWHILE
\ENDIF
\STATE  $S^{'}=$ \textsc{improve($S^{'}$)};
\STATE $\textbf{P} \gets  \textbf{P} \cup S'$;
\ENDWHILE
\STATE \textbf{return}  $\textbf{P}$
\end{algorithmic}
\end{algorithm}

\subsubsection{Crossover} \label{aa3}
The crossover procedure $\textsc{Cross}$ takes two parent solutions as input and outputs an offspring solution, as shown in Algorithm \ref{alg:cross}.  The construction process is as follows.

\begin{itemize}
  \item Randomly select two parent solutions $S_i$ and $S_j$, and let $S = S_i\cap S_j$;
  \item If $|S|=k$, then $S$ is an offspring solution; if $|S|>k$, then repeatedly delete nodes from $S$ until $|S|=k$ in a greedy way, i.e., the deleted node $v\in S$ should minimize the value of $f(S\setminus \{v\})-f(S)$; if $|S|<k$, then repeatedly select a node randomly from a large connected component and add the node to $S$, until the condition of $|S|=k$ is satisfied. {A large connected component is defined as a connected component with the number of nodes greater than $(max\_n_c + min\_n_c) / 2$, where $max\_n_c$ and $min\_n_c$ are the number of nodes in the largest and smallest connected components, respectively. In this case, the greedy strategy is employed only for deleting nodes rather than adding new ones. This is because the number of potential nodes (that are not currently part of the solution) is too large, much greater than the value of $k$. Therefore, a heuristic strategy is used instead, which involves selecting a node randomly from a large connected component to add to the solution.
}
\end{itemize}

{ In the worst case, this procedure requires greedy removal of $k$ nodes, so the time complexity is $O(k|E|)$.}

\begin{algorithm}[H]
\caption{$\textsc{Cross()}$}\label{alg:cross}
\begin{algorithmic}[1]
\STATE \textbf{Input:} Parent solutions $S_i$,$S_j$, and an integer $k$
\STATE \textbf{Output:} An offspring solution $S$

\STATE $S = S_i\cap S_j$;

\WHILE{$|S| > k$}
\STATE $v \gets \mathrm{argmin}_{x\in S}{f(S\setminus \{x\})-f(S)}$;
\STATE $S = S \setminus  \{v\}$;
\ENDWHILE

\WHILE{$|S| < k$}
\STATE $v \gets$ a random node from a large connected component in $G-S$;
\STATE $S\gets S\bigcup\{v\};$
\ENDWHILE
\STATE \textbf{return}  $S$;
\end{algorithmic}
\end{algorithm}

\subsubsection{Solution Improvement} \label{aa4}
Each offspring solution $S$ generated by  $\textsc{Cross}$ is immediately submitted to a local search procedure $\textsc{IMPROVE}$ for improvement. The procedure $\textsc{IMPROVE}$ repeatedly examines a neighbor of $S$ (by removing one node from $S$ and adding another node to $S$) to obtain a better solution, based on the observation that a cut node in a large connected component may be more important than other nodes. The pseudocode is shown in  Algorithm \ref{alg:impr}.

As shown in Algorithm \ref{alg:impr}, the termination condition is specified by the maximum number of non-improvement iterations MaxIter. That is, when the number of non-improvement iterations NonImproveIter reaches MaxIter, the local search algorithm is terminated. Then the current best solution is returned.
In each iteration, the algorithm first randomly selects a large connected component $C_R$ (line 5). {To find all the large connected components, all connected components in the graph need to be traversed. So the time complexity is $O(L)$}. If the currently best solution $S'$ is not improved in the previous $limit\_num$ iterations (i.e., $NonImproveIter > limit\_num$ in Line 6), the algorithm randomly selects a node $v\in C_R$ with the highest priority. Otherwise, when $C_R$ contains a cut node, the algorithm selects a cut node $v$ of $C_R$ (lines 7--14), and when CR does not contain a cut node, the algorithm randomly selects a node from CR (Line 10). { After a node $v$ is selected, the algorithm adds $v$ to $S$ and removes a node $u (u \neq v)$ from $S$ that minimizes the value of $f(S\setminus \{u\})$, which represents the swapping process (lines 15-17). Finally, if the new solution $S$ obtained by swapping nodes is better than the current solution $S'$, then  $S'$ is replaced by $S$ (lines 18 and 19. {Since this procedure (lines 4-19) is iterated MaxIter times, then the total time complexity is $O(MaxIter(|V| + |E|))$.}

\begin{algorithm}[!h]
\caption{ $\textsc{Improve()}$}\label{alg:impr}
\begin{algorithmic}[1]
\STATE \textbf{Input:} A solution $S$, the maximum number $MaxIter$ of iterations
\STATE \textbf{Output:} An improved solution $S^{'}$

\STATE  $S^{'} \gets S$\;

\WHILE{$NonImproveIter < MaxIter$}

\STATE $C_R\gets$ a large connected component in $G-S$\;

\IF{$NonImproveIter > limit\_num$}
    \STATE $v\gets$ a node with the highest priority in $C_R$\;
\ELSE
    \STATE $Cut\_Nodes \gets FindCutNodes(C_R)$;
    \IF{$Cut\_Nodes\neq \emptyset$}
        \STATE $v\gets$ a random node in $Cut\_Nodes$\;
    \ELSE 
       \STATE $v\gets$  a node with the highest priority in $C_R$\;
    \ENDIF
\ENDIF

\STATE $S\gets S\bigcup\{v\}$;

\STATE $u\gets \mathrm{argmin}_{x\in S, x \neq v}{f(S\setminus\{x\})}$;

\STATE $S\gets S\setminus\{u\}$;

\IF{$f(S)<f(S^{'})$}
\STATE $S^{'} \gets S$;
\STATE $NonImproveIter = 0$;
\ENDIF
\ENDWHILE

\STATE \textbf{return}  $S^{'}$
\end{algorithmic}
\end{algorithm}

\subsubsection{Population Updating} \label{aa5}
An improved individual $S^{'}$ is obtained by Algorithm 4. The current population is updated by replacing one existing individual with the improved individual $S^{'}$. The question is how to choose the replaced individual. In the population updating stage,  a strategy inspired by the population management strategies presented in \cite{7440882} is adopted to ensure the high quality of each solution in the population and to guarantee high population diversity. For a population $\textbf{P}$ with individuals $S_1,S_2,\ldots, S_{|\textbf{P}|}$,  the algorithm  uses $AD(S_i)$ to characterize the difference between an individual $S_i$ and the other individuals in $\textbf{P}$ as follows:
\begin{equation} \label{equ-difference}
AD(S_i)=\frac{1}{|\textbf{P}|} \sum\limits_{j=1}^{|\textbf{P}|}|S_i- (S_i\cap S_j)|.
\end{equation}
Based on the $AD(S_i)$ value, the following scoring rule is used to evaluate the quality of individuals in a population:
\begin{equation}\label{score}
score(S_i) = a\times drank(AD(S_i))+(1-a)\times irank(f(S_i)),
\end{equation}
where $a$ is weight of $AD(S_i)$ in the weighted sum scoring rule, $drank(AD(S_{i}))$ represents the position of $AD(S_{i})$ sorted in  descending order, and $irank(f(S_i))$ represents the position of $f(S_{i})$ sorted in ascending order for all $i$. {The time complexity of computing $AD(S_i)$ is $O(k * |\textbf{P}|)$}.

The updating procedure is shown in Algorithm \ref{alg:upda}  where the individual $S_i$ with the largest score in the current population $\textbf{P}$ is selected and replaced with the improved individual $S^{'}$.

\begin{algorithm}[H]
\caption{$\textsc{UpdatePoP()}$}\label{alg:upda}
\begin{algorithmic}[1]
\STATE \textbf{Input:} A population $\textbf{P}$ and an improved solution $S'$
\STATE \textbf{Output:} An updated population $\textbf{P}$

\STATE $S \gets $ an individual in $\textbf{P}$ with the largest score value;
\STATE $\textbf{P} \gets (\textbf{P}\cup S')\setminus S$;
\STATE \textbf{return}  $\textbf{P}$
\end{algorithmic}
\end{algorithm}

{ 

As we have already explained in this section, the time complexity of each procedure is as follows: $O(k|E|)$ for the worst case of \textsc{cross}, and $O(MaxIter(|V| + |E|))$ for \textsc{improve}. In \textsc{UpdatePoP}, the time complexity of computing  $AD(S_i)$ values is $O(k * |\textbf{P}|^2)$. Since the values of $k$ and  $|\textbf{P}|$ are normally constant numbers, the time complexity of the genetic algorithm part is $O(MaxIter(|V| + |E|)$. Note that the neural network is pretrained offline. Once trained, it can be used to handle all instances. So the training time of the neural network is not included in the computation time of the proposed algorithm.
Hence, the total time complexity of our algorithm is $O(MaxIter(|V| + |E|))$.}

\section{Experimental Evaluation}\label{sec-4}

This section presents the performance evaluation results of the proposed K2GA algorithm compared to two state-of-the-art algorithms.

\subsection{Benchmark Datasets and Training Datasets}
The benchmark dataset used to evaluate the effectiveness of K2GA  is a most commonly used benchmark \cite{zhou2018memetic}, which comprises 26 real-world instances\footnote{Available at http://www.di.unito.it/aringhie/cnp.html.}.
These instances \cite{aringhieri2016general} are from a variety of fields and have different numbers of nodes (ranging from 121 to 23133), and edge counts (ranging from 190 to 198050). Some examples include biology (such as cattle-related protein interaction network,  $Bovine$; interaction graphs between bacteria and E. coli, $EColi$), electronics (like the power distribution network graph in the United States, $Circuit$), and transportation (such as the main flight connection in the United States in 1997, $USAir97$, and the train network around Rome,$TrainsRome$), as well as other various complex networks.

For pretraining a general neural network for solving various graphs, 300\footnote{We have performed experiments to determine the minimum number of graphs required for a well-trained neural network model. Our findings indicate that using 300  graphs is more suitable. Additional information can be found in the supplementary material.}  graphs are generated using the NetworkX  \cite{SciPyProceedings_11} random graph generator. These randomly generated graphs consisted of 100-300 nodes with node degrees ranging from 1 to 63 and edge counts ranging from 54 to 6360. Out of these 300 graphs, 60\% were designated for the training set, 20\% for the validation set, and the remaining 20\% for the test set.

\subsection{Experimental Settings}

All experiments were performed on a Dell computer with a 2.4-GHz Intel(R) Core(TM) i9-10885H processor, 32 GB of RAM, and a Quadro RTX 4000 GPU.

The  K2GA algorithm, like most of the heuristic algorithms for solving the CNP-1a problem, involves hyperparameters. Regarding the neural network part, six hyperparameters were tuned:  $weight\_decay$ , $learning\_rate$ , the number of epochs,   $hidden\_dims$ (the number of hidden layer neurons), $layer\_num$ (the number of hidden layers), and $aggregator\_type$ (GAT neighborhood aggregation schemes).

To determine suitable hyperparameters, a random search method was employed. Each group of parameters was trained five times, resulting in a total of 300 searches. During the training process, the group of hyperparameters and corresponding models with the highest area under the curve (AUC) value were saved. The hyperparameter values are shown in Table \ref{tab-2}.

For the genetic algorithm, the required parameters include $|\textbf{P}|$ (the population size), $p$ (the probability selected for the crossover operation), $a$ (the proportion of similarity degree during updating population), {\it cutoff} (the cutoff running time of the algorithm), $MaxIter$ (the maximum number of rounds during the local search that ensures the solution is not improved), and $limit\_num$ (the maximum number of rounds that ensures the solution is not improved when selecting a cut node).
Among these parameters, $|\textbf{P}|$, $p$, $a$, and {\it cutoff} were set according to \cite{zhou2018memetic}. The other two parameters (i.e., $MaxIter$ and $limit\_num$) were tuned in a trial-and-error manner. Table \ref{tab-3} lists the values of the hyperparameters.

\begin{table}[H]
\caption{Parameter settings of the neural network \label{tab-2}}
\centering
\begin{tabular}{|c|c|c|}
\hline
Hyperparameters & description & value\\
\hline
 $weight\_decay$ & weight decay & 0.001\\
\hline
$learning\_rate$   & learning rate & 0.001\\
\hline
$epoch$  & the number of epochs & 200\\
\hline
$hidden\_dims$  & the number of hidden layer nodes & 64\\
\hline
$layer\_num$  & the number of hidden layers & 4\\
\hline
$aggregator\_type$  &neighborhood aggregation schemes & mean\\
\hline
\end{tabular}
\end{table}

\begin{table}[H]
\caption{Parameter settings of the genetic algorithm \label{tab-3}}
\centering
\begin{tabular}{|c|c|c|}
\hline
Parameters & description & value\\
\hline
 $|\textbf{P}|$ & population size & 20\\
\hline
$p$   & crossover probability & 0.9\\
\hline
$a$ & proportion of similarity & 0.6\\
\hline
{\it cutoff}  & cutoff running time & 3600(s)\\
\hline
{\it MaxIter}  & unimproved rounds in local search & 1500\\
\hline
$limit\_num$  & unimproved rounds by a cut node & 100\\
\hline
\end{tabular}
\end{table}
\subsection{Baselines and Metrics}

The proposed K2GA algorithm was compared with two state-of-the-art heuristic algorithms for solving CNP-1a, MACNP \cite{zhou2018memetic} and VPMS \cite{zhou2020variable}. The time complexity of K2GA is at the same level as both algorithms. Each algorithm was run five times in the same environment. The executable programs of the MACNP and VPMS algorithms were provided by the authors of each paper \cite{zhou2018memetic,zhou2020variable}. To evaluate the performance of the algorithms, we use the standard metrics that are commonly employed in the study for solving CNPs. There metrics include the best objective value ($f^*$), the median objective value ($f_m$), and the average objective value ($\overline{f}$) obtained after five runs. Lower values for these metrics indicate higher algorithm performance.

We use the widely recognized Wilcoxon signed rank test\cite{10.5555/1248547.1248548} to compare the results and determine the significance of differences in each comparison indicator between a baseline algorithm and K2GA. In this test, if the p-value of algorithm X is equal to or less than 0.05  (in case we assume a significance level of 5\%), it is considered that K2GA is significantly better than X. 

\subsection{Comparison Experiments on Real-world Dataset}\label{bb2}
Tables  \ref{tab-4} show the experimental results of the three algorithms for the 26 real-world benchmark instances. The first column lists the name of instances (Instance), while the second column provides the known best value (KBV) reported in  \cite{zhou2018memetic, zhou2020variable}, and columns 3-5 show the results of the MACNP algorithm, including the best objective value $f^*$ obtained after five runs, the median objective value $f_m$, and the average objective value $\overline{f}$. Similarly, columns 6-9 and 10-12 report the results of VPMS and K2GA, respectively.  The best solutions obtained by the three algorithms are written in bold in Tables \ref{tab-4}. In cases where K2GA obtained better results than KBV,  the corresponding result of K2GA is marked by `*' in the superscript. The last row in Table \ref{tab-4} shows the results on p-values.

\begin{table*}[hbtp]
 \centering
	\scriptsize
	\renewcommand\tabcolsep{6pt}
	\renewcommand{\arraystretch}{1.5}
	\caption{COMPARATIVE PERFORMANCE OF THE PROPOSED K2GA WITH MACNP AND VPMS ON REAL-WORLD BENCHMARK}
 \begin{threeparttable}
\begin{tabular}{c|c|ccc|ccc|ccc}
\hline
\multirow{2}{*}{Instances} & \multirow{2}{*}{KBV} & \multicolumn{3}{c|}{MACNP}                            & \multicolumn{3}{c|}{VPMS\tnote{1}}                        & \multicolumn{3}{c}{K2GA}                               \\ \cline{3-11}
                   &                    & $f^*$           & $f_m$            & $\overline{f}$    & $f^*$          & $f_m$           & $\overline{f}$ & $f^*$               & $f_m$            & $\overline{f}$  \\ \hline
Bovine      & 268      & \textbf{268}  & \textbf{268}  & \textbf{268}  & \textbf{268}   & \textbf{268}   & \textbf{268}      & \textbf{268}      & \textbf{268}      & \textbf{268}        \\
Circuit     & 2099     & \textbf{2099} & \textbf{2099} & \textbf{2099} & \textbf{2099}  & \textbf{2099}  & \textbf{2099}     & \textbf{2099}     & \textbf{2099}     & \textbf{2099}       \\
Ecoli       & 806      & \textbf{806}  & \textbf{806}  & \textbf{806}  & \textbf{806}   & \textbf{806}   & \textbf{8806}     & \textbf{8806}     & \textbf{8806}     & \textbf{8806}       \\
humanDi     & 1115     & \textbf{1115} & \textbf{1115} & \textbf{1115} & \textbf{1115}  & \textbf{1115}  & \textbf{1115}     & \textbf{1115}     & \textbf{1115}     & \textbf{1115}       \\
Treni\_R    & 918      & \textbf{918}  & \textbf{918}  & \textbf{918}  & \textbf{918}   & \textbf{918}   & \textbf{918}      & \textbf{918}      & \textbf{918}      & \textbf{918}        \\
yeast1      & 1412     & \textbf{1412} & \textbf{1412} & \textbf{1412} & \textbf{1412}  & \textbf{1412}  & \textbf{1412}     & \textbf{1412}     & \textbf{1412}     & \textbf{1412}       \\
astroph     & 53963375 & 59743097      & 61326804      & 61017141.0    & 58322396       &                & 59563941.1        & \textbf{54299368} & \textbf{55517566} & \textbf{55587112.2} \\
condmat     & 2298596  & 8890056       & 9117119       & 9183903       & 6843993        &                & 7813436.7         & \textbf{3246734}  & \textbf{3425268}  & \textbf{3446396.6}  \\
EU\_flights & 348268   & 350762        & 356631        & 354283.4      & 348268         & 350762         & 350263.2          & \textbf{349937}   & \textbf{349937}   & \textbf{349937}     \\
facebook    & 420334   & 729312        & 768878        & 768983.0      & 701838         & 773723         & 757912.8          & \textbf{680273}   & \textbf{685603}   & \textbf{684822.4}   \\
grqc        & 13596    & 13630         & 13645         & 13645.2       & 13634          & 13643          & 13648.2           & \textbf{13620}    & \textbf{13621}    & \textbf{13630}      \\
H1000       & 306349   & 310131        & 311376        & 311443.4      & 310377         & 311909         & {311864.2} & \textbf{306349}   & \textbf{308538}   & \textbf{309082.6}   \\
H2000       & 1242739  & 1262189       & 1276169       & 1274680.4     & 1243408        & 1249950        & {1252515}  & \textbf{1241508$^*$}  & \textbf{1247873}  & \textbf{1248012.4}  \\
H3000a      & 2840690  & 2925336       & 2934076       & 2942835.4     & 2841068        & 2855375        & {2857652}  & \textbf{2787564$^*$}  & \textbf{2827590}  & \textbf{2821190.2}  \\
H3000b      & 2837584  & 2897868       & 2964454       & 2960600.8     & 2846706        & 2877323        & {2870202}  & \textbf{2807504$^*$}  & \textbf{2821153}  & \textbf{2828265}    \\
H3000c      & 2835369  & 2870210       & 2942813       & 2960441.4     & 2837604        & 2838083        & {2844788}  & \textbf{2783732$^*$}  & \textbf{2809488}  & \textbf{2806253}    \\
H3000d      & 2828492  & 2922443       & 2944055       & 2970674.1     & 2829042        & 2831268        & {2840237}  & \textbf{2804245$^*$}  & \textbf{2815022}  & \textbf{2819274}    \\
H3000e      & 2843000  & 2910890       & 2915637       & 2926384.2     & 2845657        & 2854896        & {2862952}  & \textbf{2246267$^*$}  & \textbf{2281435}  & \textbf{2278445}    \\
Oclink       & 5038611  & 5285856       & 5393573       & 5356226.6     & 5120498        & 5137544        & {5135622}  & \textbf{5008252$^*$}  & \textbf{5056719}  & \textbf{5057784}    \\
H5000       & 7964765  & 8458059       & 8518529       & 8506245.0     & 8103176        & 8160400        & {8151215}  & \textbf{8000484}  & \textbf{8022376}  & \textbf{8019440}    \\
hepph       & 6155877  & 9843207       & 10142454      & 10176952.4    & 10400613       & 10584783       & {10623310} & \textbf{6844217}  & \textbf{7103269}  & \textbf{7114137}    \\
hepth       & 106276   & 106596        & 108048        & 107975.6      & 115622         & 116231         & {117638.2} & \textbf{105396$^*$}   & \textbf{107092}   & \textbf{106887.8}   \\
OClinks     & 611253   & 614468        & 615575        & 615574.2      & 614467        & \textbf{614469}          & 615131.6          & \textbf{611253}               & 614494        & \textbf{614496.6}                    \\
openflights & 26783    & 28700         & 28920         & 28908.2       & \textbf{26874} & \textbf{26875} & \textbf{27685.4}  & 28700             & 28834             & 28807.2             \\
powergrid   & 15862    & 15882         & 15934         & 15927.8       & 15968          & 16011          & 16000.8           & \textbf{15881}    & \textbf{15910}    & \textbf{15906}      \\
USAir97     & 4336     & \textbf{4336} & \textbf{4336} & \textbf{4336} & 4726           & 5444           & 5156.8            & 5418              & 5444              & 5433.6         \\ \hline
p-value    & -         & 5.1e-4        & 4.7e-4         &4.7e-4        &1.0e-3          &2.1e-3           &3.3e-4                  &-                   &-                &- \\ \hline
\end{tabular}
 \begin{tablenotes}    
        \footnotesize               
        \item[1]  Due to errors in the executable program for astroph and condmat instances, we use the $f^*$ and $\overline{f}$ provided in the paper \cite{zhou2020variable}. However, the $f_m$ values for these two instances are not provided.     
      \end{tablenotes}            
    \end{threeparttable}       
\label{tab-4}%
\end{table*}

According to the results, the K2GA algorithm outperformed the other two algorithms in most of the testing instances. Specifically, it achieved best values on both $f^*$ and $\overline{f}$ for 23 out of the 26 instances, including 8 new upper bounds. Compared to the MACNP algorithm, K2GA performed better on 18 instances regarding both $f^*$ and $\overline{f}$. Similarly, compared to the VPMS algorithm, K2GA performed better on 18 instances regarding both $f^*$ and $\overline{f}$.

Regarding $f^*$, $f_m$, and $\overline{f}$, K2GA performed better than MACNP and VPMS on 18 instances, respectively. Although K2GA performed slightly worse than  MACNP regarding $f^*$ on the powergrid instance, it outperformed MACNP on both $f_m$ and $\overline{f}$. Additionally, the K2GA algorithm performed worse than VPMS  on only two instances, EU\_flights and USAir97. The above results indicated that the proposed K2GA algorithm could be highly competitive for solving the CNP-1a problem on real-world data compared to the state-of-the-art algorithms.Since p-values are significantly lower than the significance level of 0.05 in terms of the metrics $f^*$, $f_m$, and $\overline{f}$, there is a significant difference between the performance of the baselines and that of K2GA. This indicates that K2GA outperforms MACNP and VPMS significantly.

\subsection{Performance verification with limited running time}

In previous experiments, we have shown that K2GA outperforms the baselines when given a runtime of 3600s,  which is commonly used in the literature on CNP-1a. To assess the performance of K2GA with a limited runtime, we conduct experiments on the best solutions obtained by K2GA, MACNP, and VPMS, when the running time is less than 3600s. We set different runtime limits from 120s to 3600s with an increase of 120s, and obtained 30 data points for each algorithm on each instance. Then, we fit these 30 data points of the running results into a curve. As the real-world dataset includes graph instances with a number of nodes ranging from 100 to 20,000, we showcase the results of the algorithms on several representative instances including Oclink, H2000, H5000, Facebook, hepth, and Hepph, in Figures \ref{Oclink}-\ref{hepph}. These instances include 1,000, 2,000, 4,039, 5,000, 9,877, and 12,008 nodes, respectively. The X-axis in these figures represents the running time of each algorithm, and the Y-axis represents the best objective values obtained by the algorithms. The lower the curve, the better the algorithm's performance.  It can be seen from these figures that fluctuations can occur due to randomness, but the overall trend of the curves are decreasing as the running time increases. The curves of K2GA are lower than those of MACNP and VPMS, indicating that K2GA outperforms MACNP and VPMS even with limited runtime.

\begin{figure*}[htbp]
\centering
\begin{minipage}[t]{0.48\textwidth}
\centering
\includegraphics[width=7.5cm]{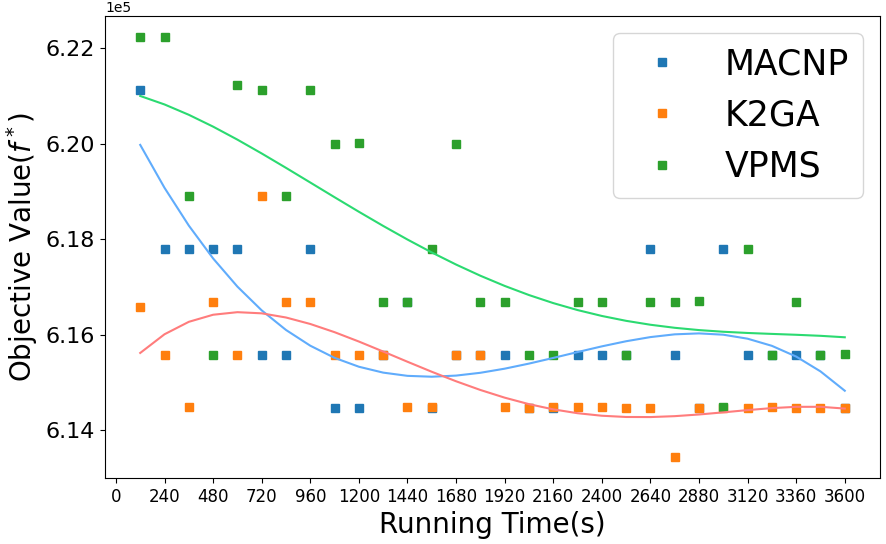}
\caption{Performance on Oclink with limited running time}
\label{Oclink}
\end{minipage}
\begin{minipage}[t]{0.48\textwidth}
\centering
\includegraphics[width=7.5cm]{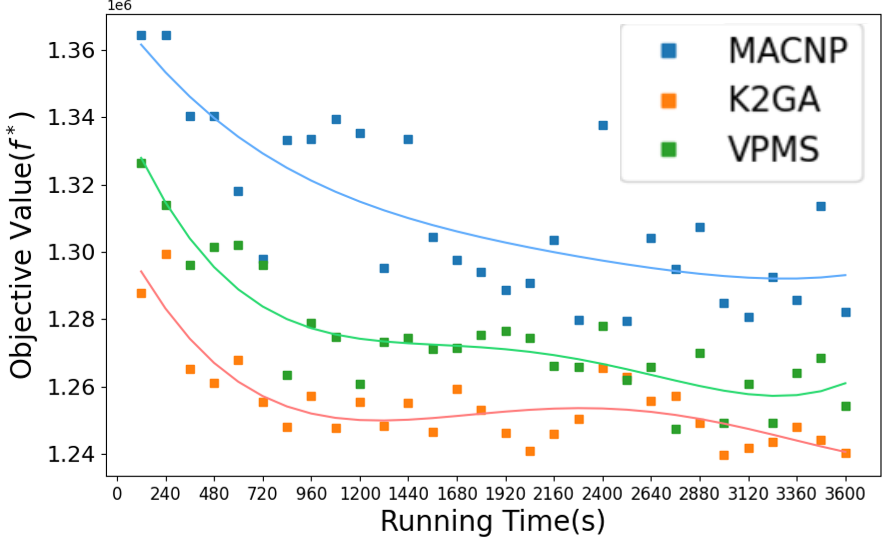}
\caption{Performance on H2000 with limited running time}
\label{H2000}
\end{minipage}
\quad

\vspace{2ex}

\begin{minipage}[t]{0.48\textwidth}
\centering
\includegraphics[width=7.5cm]{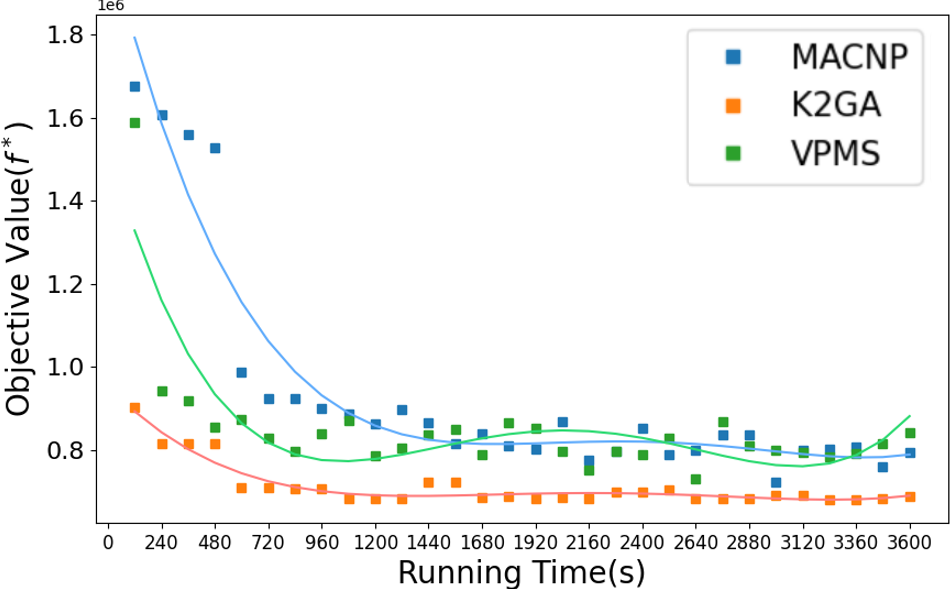}
\caption{Performance on facebook with limited running time}
\label{facebook}
\end{minipage}
\begin{minipage}[t]{0.48\textwidth}
\centering
\includegraphics[width=7.5cm]{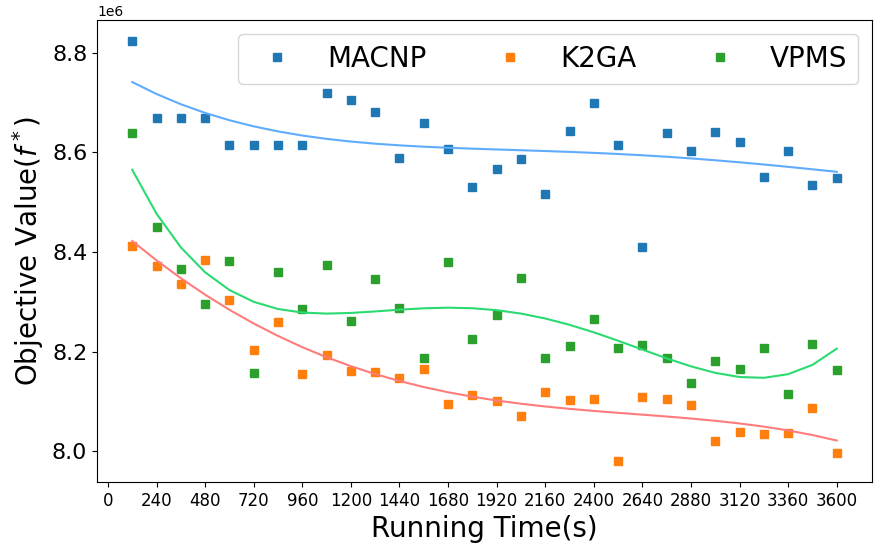}
\caption{Performance on  H5000 with limited running time}
\label{H5000}
\end{minipage}
\quad

\vspace{2ex}

\begin{minipage}[t]{0.48\textwidth}
\centering
\includegraphics[width=7.5cm]{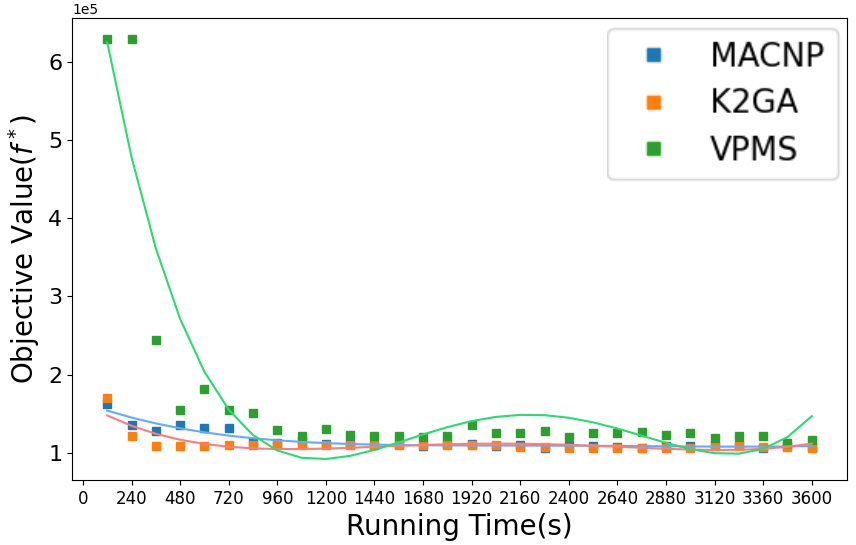}
\caption{Performance on hepth with limited running time}
\label{hepth}
\end{minipage}
\begin{minipage}[t]{0.48\textwidth}
\centering
\includegraphics[width=7.5cm]{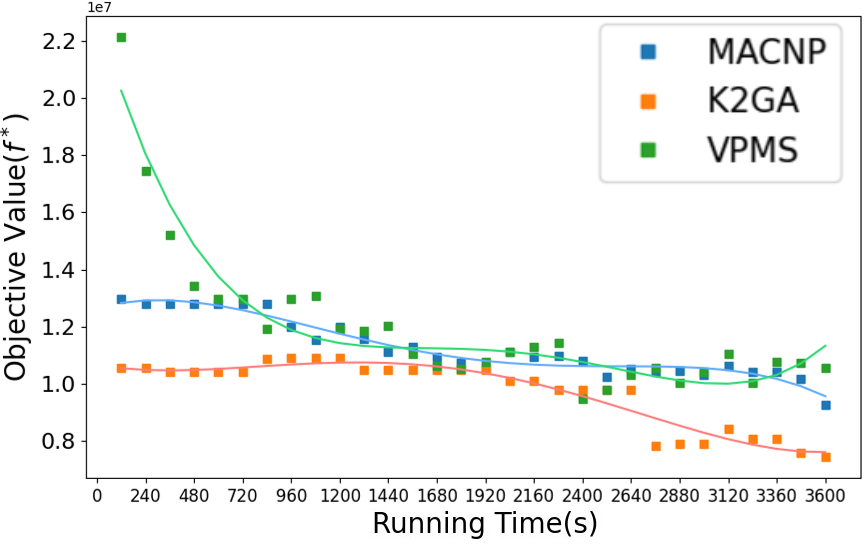}
\caption{Performance on  hepph with limited running time}
\label{hepph}
\end{minipage}

\end{figure*}

\subsection{Effectiveness of the K2GA Neural Network}
In this section, we evaluate the effectiveness of K2GA  by comparing it with its two variants: K2GA\_randInit and K2GA\_randInit\_longImprove.  K2GA\_randInit uses random nodes in  \textsc{InitialPoP} to generate the initial solutions. K2GA\_randInit\_longImprove is designed based on K2GA\_randInit by giving the improvement of initial solutions (Algorithm \ref{alg:impr}) in the population  under a longer time limit. Since it takes approximately 24 hours to train the neural network, this time limit is also set as 24 hours.

We use a metric called Comparative Advantage (CA) to compare the performance of these three algorithms. The CA is defined as $(C_{X} - C)/C * 100$, where $C_{X}$ is the $f^*$ value obtained by an algorithm $X$ ($X$ can be K2GA, K2GA\_randInit or K2GA\_randInit\_longImprove), and $C$ is the best $f^*$ value achieved among the three algorithms. The lower the CA value, the better the algorithm performs.  The comparison results are presented in Figure \ref{CA}. The X-axis represents different instances, and the Y-axis represents the corresponding CA values. It can be seen from this figure that K2GA's overall performance is superior to the other two variants.

We also conducted experiments to test the effectiveness of the neural network model on the initial solution. We compared the average objective value $\overline{f}$ with different initial solutions, where the neural network predicted nodes accounted for 20\%, 40\%, 60\%, 80\%, and 100\% of the total nodes. When the percentage of predicted nodes was 0\%, the initial solution was generated randomly, as discussed earlier. We repeated each case of the experiment five times and calculated the average $\overline{f}$ value. We recorded the difference between $\overline{f}$ and the average objective value in the case of a random initialization. The results are shown in Figures \ref{fig-2}--\ref{fig-5}. The Y-axis represents the difference in average objective values, and the X-axis shows the percentage of GNN-predicted nodes in the initial solution. As expected, the quality of the initial solution improved as we increased the percentage of GNN-predicted nodes in the initial solution.


\begin{figure}[!t]
\centering
\includegraphics[width=8.5cm]{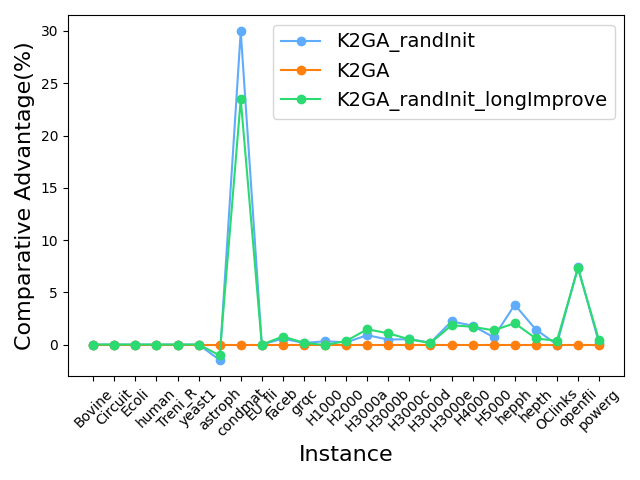}
\caption{Comparative Advantage on Realworld Instances}
\label{CA}
\end{figure}

\begin{figure*}[htbp]
\centering
\begin{minipage}[t]{0.48\textwidth}
\centering
\includegraphics[width=5.8cm]{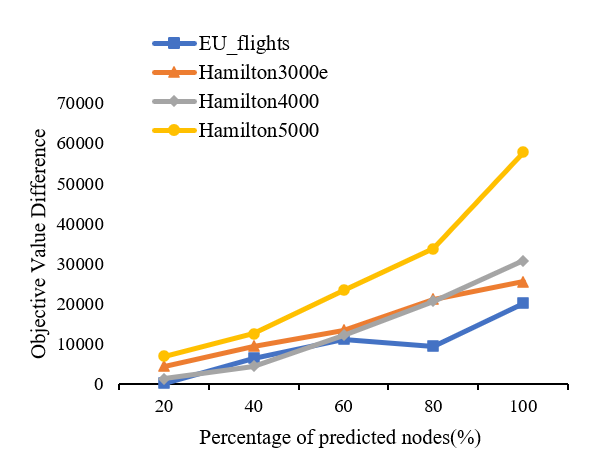}
\caption{Objective value difference in the range of 10,000-60,000}
\label{fig-2}
\end{minipage}
\begin{minipage}[t]{0.48\textwidth}
\centering
\includegraphics[width=5.8cm]{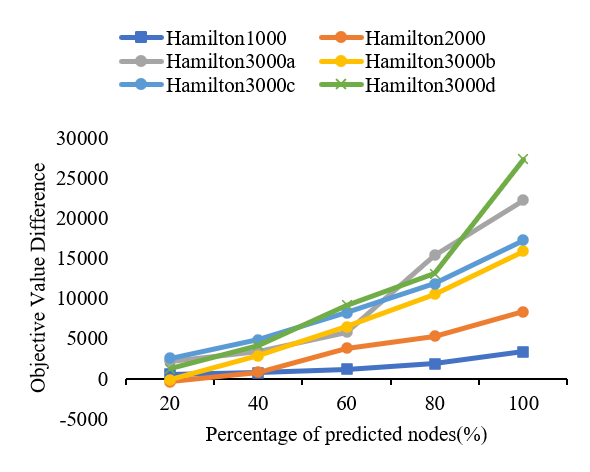}
\caption{Objective value difference in the range of 5,000-30,000}
\label{fig-3}
\end{minipage}
\end{figure*}

\begin{figure*}[htbp]
\begin{minipage}[t]{0.48\textwidth}
\centering
\includegraphics[width=5.8cm]{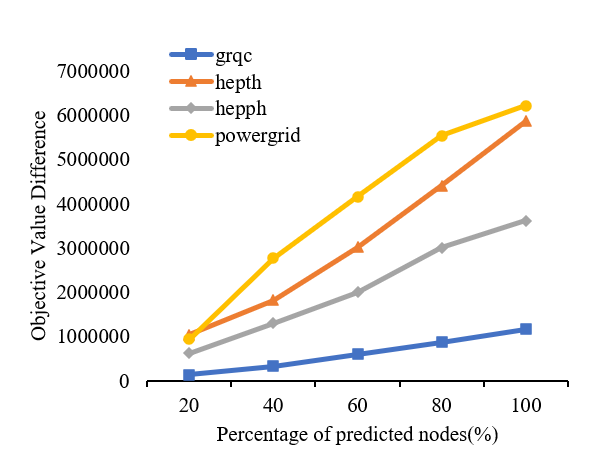}
\caption{Objective value difference in the range of 1,000,000-7,000,000}
\label{fig-4}
\end{minipage}
\begin{minipage}[t]{0.48\textwidth}
\centering
\includegraphics[width=5.8cm]{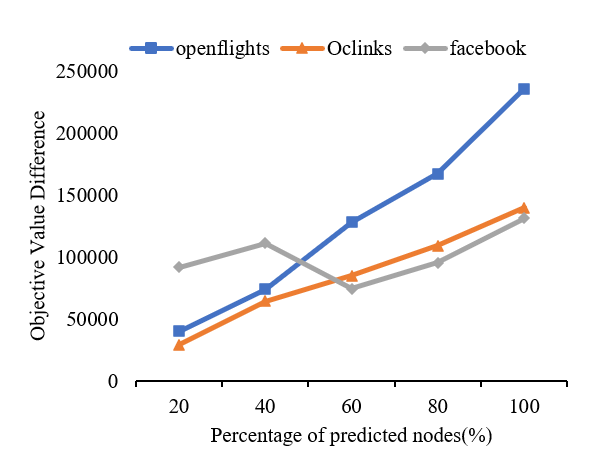}
\caption{Objective value difference in the range of 50,000-250,000}
\label{fig-5}
\end{minipage}

\end{figure*}

\subsection{Effectiveness of the Cut node-Based Greedy Strategy}

In the K2GA algorithm, the \textsc{improve} function uses a cut node-based greedy strategy (line 10). To determine the effectiveness of this strategy, a variant of K2GA called K2GA\_no\_cutnode was created. This variant drops the cut node-based greedy strategy and instead selects the node with the largest priority when adding nodes to a solution. Table \ref{tab-6} compares the results of K2GA and K2GA\_no\_cutnode. On most instances, the best and average values obtained by K2GA are better than those obtained by K2GA\_no\_cutnode.

\begin{table}[hbtp]
 \centering
	\scriptsize
	\renewcommand\tabcolsep{7pt}
	\renewcommand{\arraystretch}{1.5}
	\caption{COMPARATIVE PERFORMANCE OF K2GA WITH K2GA\_no\_cutnode ON REAL-WORLD BENCHMARK}
 \begin{threeparttable}
\begin{tabular}{c|cc|cc}
\hline
\multirow{2}{*}{Instances}  & \multicolumn{2}{c|}{K2GA\_no\_cutnode} & \multicolumn{2}{c}{K2GA} \\ \cline{2-5}
                                & $f^*$                    & $\overline{f}$    & $f^*$                  & $\overline{f}$   \\ \hline

astroph       & 58664411 & 59124601.2       & \textbf{54299368} & \textbf{55587112.2} \\
condmat       & 3539445  & 3670018.0        & \textbf{3246734}  & \textbf{3446396.6}  \\
EU\_flights   & \textbf{348268}   & \textbf{349603.2}         & 349937            & 349937              \\
facebook      & 705385   & 721584.2         & \textbf{680273}   & \textbf{684822.4}   \\
grqc          & 13719    & 13756.8          & \textbf{13620}    & \textbf{13630}      \\
Hamilton1000  & 307804   & 309810.6         & \textbf{306349}   & \textbf{309082.6}   \\
Hamilton2000  & 1247021  & 1253408.2        & \textbf{1241508}  & \textbf{1248012.4}  \\
Hamilton3000a & 2816498  & 2835387.2        & \textbf{2787564}  & \textbf{2821190.2}  \\
Hamilton3000b & 2841577  & 2845781.4        & \textbf{2807504}  & \textbf{2828265}    \\
Hamilton3000c & 2809612  & 2829671.6        & \textbf{2783732}  & \textbf{2806253}    \\
Hamilton3000d & 2839413  & 2859510.8        & \textbf{2804245}  & \textbf{2819274}    \\
Hamilton3000e & 2299723  & 2311729.0        & \textbf{2246267}  & \textbf{2278445}    \\
Hamilton4000  & 5094115  & 5149465.4        & \textbf{5008252}  & \textbf{5057784}    \\
Hamilton5000  & 8087442  & 8100408.0        & \textbf{8000484}  & \textbf{8019440}    \\
hepph         & 7795002  & 8034759.0        & \textbf{6844217}  & \textbf{7114137}    \\
hepth         & 107715   & 108162.8         & \textbf{105396}   & \textbf{106887.8}   \\
OClinks       & 614467   & 614911.2         & \textbf{611253}   & \textbf{614496.6}   \\
openflights   & 26874    & \textbf{27401.2} & \textbf{26842}    & 28435.6             \\
powergrid     & 15890    & 15920.8          & \textbf{15881}    & \textbf{15906}    \\ \hline
\end{tabular}

    \end{threeparttable}       
\label{tab-6}%
\end{table}

To summarize, the experimental results indicate that the K2GA algorithm proposed in this study outperformed two state-of-the-art algorithms for the CNP-1a problem and achieved eight new upper bounds on the best-known objective value. The ablation experiments conducted on real-world instances further demonstrate the effectiveness of the key strategies proposed in K2GA.

\section{Conclusion}\label{sec-5}

In this paper, we presented a graph neural network (GNN)-based knowledge-guided genetic algorithm called K2GA, which is based on a genetic algorithm and employs a graph neural network to guide the search for critical nodes in a graph. Specifically, the neural network is used to predict the optimal set of critical nodes that can be used to create a high-quality initial population for the genetic algorithm. In the second phase, the genetic algorithm searches for the optimal solution using the initial population generated by the GNN. The K2GA algorithm has been tested against two state-of-the-art heuristic algorithms in solving the CNP-1a problem on 26 instances. The experiment results indicate that K2GA outperformed the other algorithms on the test instances. Furthermore, K2GA has improved the best upper bounds on the best objective values for eight real-world instances. The effectiveness of the K2GA neural network has also been verified by comparing it with its variants. These findings suggest that the K2GA algorithm is a promising approach for solving CNP-1a, and the GNN-based approach can be used to guide the search for critical nodes in various potential applications.


This work opens an avenue for solving the CNP-1a problem via a knowledge-guided genetic algorithm. The proposed strategy (i.e., deep learning-based initialization for genetic algorithms ) can be adapted to other CNPs, although the neural networks have to be designed accordingly. In the current approach, the training data are generated by running a genetic algorithm multiple times on the same graph. The union of the obtained solutions (i.e., selected nodes) is used as "labels". However, the frequency of nodes included in the solutions may vary, and the solution quality may differ in each run. The current "labels" are generated without considering these differences, which would be regarded as future work. Since CNP-1a is a subset selection problem with the cardinality condition, one possible direction is to formulate it as a two-objective problem and solve it using multi-objective algorithms such as GSEMO \cite{LIU2023}. Additionally, other knowledge-guided genetic frameworks will be explored for solving CNPs or other $\mathcal{NP}$-hard problems. This will offer valuable insights into solving complex optimization problems and developing more efficient and effective algorithms in the future.

\newpage

\begin{thebibliography}{10}
\providecommand{\url}[1]{#1}
\csname url@samestyle\endcsname
\providecommand{\newblock}{\relax}
\providecommand{\bibinfo}[2]{#2}
\providecommand{\BIBentrySTDinterwordspacing}{\spaceskip=0pt\relax}
\providecommand{\BIBentryALTinterwordstretchfactor}{4}
\providecommand{\BIBentryALTinterwordspacing}{\spaceskip=\fontdimen2\font plus
\BIBentryALTinterwordstretchfactor\fontdimen3\font minus \fontdimen4\font\relax}
\providecommand{\BIBforeignlanguage}[2]{{%
\expandafter\ifx\csname l@#1\endcsname\relax
\typeout{** WARNING: IEEEtran.bst: No hyphenation pattern has been}%
\typeout{** loaded for the language `#1'. Using the pattern for}%
\typeout{** the default language instead.}%
\else
\language=\csname l@#1\endcsname
\fi
#2}}
\providecommand{\BIBdecl}{\relax}
\BIBdecl

\bibitem{1998Dynamics}
Y.~Bar-Yam, S.~R. Mckay, and W.~Christian, ``Dynamics of complex systems (studies in nonlinearity),'' \emph{Computers in Physics}, vol.~12, no.~4, pp. 335--336, 1998.

\bibitem{2012Interconnectedness}
A.~Zelenkauskaite, N.~Bessis, S.~Sotiriadis, and E.~Asimakopoulou, ``Interconnectedness of complex systems of internet of things through social network analysis for disaster management,'' in \emph{2012 4th International Conference on Intelligent Networking and Collaborative Systems (INCoS)}, 2012, pp. 503--508.

\bibitem{2014A}
M.~Ventresca and D.~Aleman, ``A derandomized approximation algorithm for the critical node detection problem,'' \emph{Computers \& Operations Research}, vol.~43, no. MAR., pp. 261--270, 2014.

\bibitem{WangW22}
J.~Wang and L.~Wang, ``A cooperative memetic algorithm with learning-based agent for energy-aware distributed hybrid flow-shop scheduling,'' \emph{{IEEE} Trans. Evol. Comput.}, vol.~26, no.~3, pp. 461--475, 2022.

\bibitem{2006Identifying}
S.~P. Borgatti, ``Identifying sets of key players in a social network,'' \emph{Computational \& Mathematical Organization Theory}, vol.~12, no.~1, pp. 21--34, 2006.

\bibitem{00030HLS21}
K.~Wu, J.~Liu, X.~Hao, P.~Liu, and F.~Shen, ``An evolutionary multiobjective framework for complex network reconstruction using community structure,'' \emph{{IEEE} Trans. Evol. Comput.}, vol.~25, no.~2, pp. 247--261, 2021.

\bibitem{1982Most}
H.~W. Corley and D.~Y. Sha, ``Most vital links and nodes in weighted networks,'' \emph{Operations Research Letters}, vol.~1, no.~4, pp. 157--160, 1982.

\bibitem{ZHU2024127195}
E.~Zhu, H.~Wang, Y.~Zhang, K.~Zhang, and C.~Liu, ``Phee: Identifying influential nodes in social networks with a phased evaluation-enhanced search,'' \emph{Neurocomputing}, vol. 572, p. 127195, 2024.

\bibitem{9434427}
L.~Wang, L.~Ma, C.~Wang, N.-G. Xie, J.~M. Koh, and K.~H. Cheong, ``Identifying influential spreaders in social networks through discrete moth-flame optimization,'' \emph{IEEE Transactions on Evolutionary Computation}, vol.~25, no.~6, pp. 1091--1102, 2021.

\bibitem{2011Finding}
C.~T. Li, S.~D. Lin, and M.~K. Shan, ``Finding influential mediators in social networks,'' in \emph{Proceedings of the 20th International Conference on World Wide Web, WWW 2011, Hyderabad, India, March 28 - April 1, 2011 (Companion Volume)}, 2011, pp. 75--76.

\bibitem{LIU2023119140}
C.~Liu, S.~Ge, and Y.~Zhang, ``Identifying the cardinality-constrained critical nodes with a hybrid evolutionary algorithm,'' \emph{Information Sciences}, vol. 642, p. 119140, 2023.

\bibitem{jhoti2007structure}
H.~Jhoti and A.~R. Leach, \emph{Structure-based drug discovery}.\hskip 1em plus 0.5em minus 0.4em\relax Springer, 2007, vol.~1.

\bibitem{stromgaard2009textbook}
K.~Stromgaard, P.~Krogsgaard-Larsen, and U.~Madsen, \emph{Textbook of drug design and discovery}.\hskip 1em plus 0.5em minus 0.4em\relax CRC press, 2009.

\bibitem{tomaino2012studying}
V.~Tomaino, A.~Arulselvan, P.~Veltri, and P.~M. Pardalos, ``Studying connectivity properties in human protein--protein interaction network in cancer pathway,'' in \emph{Data Mining for Biomarker Discovery}.\hskip 1em plus 0.5em minus 0.4em\relax Springer, 2012, pp. 187--197.

\bibitem{tao2006epidemic}
Z.~Tao, F.~Zhongqian, and W.~Binghong, ``Epidemic dynamics on complex networks,'' \emph{Progress in Natural Science}, vol.~16, no.~5, pp. 452--457, 2006.

\bibitem{687888}
F.~Zhang, Y.~Zhang, and A.~Nee, ``Using genetic algorithms in process planning for job shop machining,'' \emph{IEEE Transactions on Evolutionary Computation}, vol.~1, no.~4, pp. 278--289, 1997.

\bibitem{zhou2022graph}
Y.~Zhou, H.~Zheng, X.~Huang, S.~Hao, D.~Li, and J.~Zhao, ``Graph neural networks: Taxonomy, advances, and trends,'' \emph{ACM Transactions on Intelligent Systems and Technology (TIST)}, vol.~13, no.~1, pp. 1--54, 2022.

\bibitem{arulselvan2009detecting}
A.~Arulselvan, C.~W. Commander, L.~Elefteriadou, and P.~M. Pardalos, ``Detecting critical nodes in sparse graphs,'' \emph{Computers \& Operations Research}, vol.~36, no.~7, pp. 2193--2200, 2009.

\bibitem{ventresca2014fast}
M.~Ventresca and D.~Aleman, ``A fast greedy algorithm for the critical node detection problem,'' in \emph{International Conference on Combinatorial Optimization and Applications}.\hskip 1em plus 0.5em minus 0.4em\relax Springer, 2014, pp. 603--612.

\bibitem{pullan2015heuristic}
W.~Pullan, ``Heuristic identification of critical nodes in sparse real-world graphs,'' \emph{Journal of Heuristics}, vol.~21, no.~5, pp. 577--598, 2015.

\bibitem{addis2016hybrid}
B.~Addis, R.~Aringhieri, A.~Grosso, and P.~Hosteins, ``Hybrid constructive heuristics for the critical node problem,'' \emph{Annals of Operations Research}, vol. 238, no.~1, pp. 637--649, 2016.

\bibitem{aringhieri2016local}
R.~Aringhieri, A.~Grosso, P.~Hosteins, and R.~Scatamacchia, ``Local search metaheuristics for the critical node problem,'' \emph{Networks}, vol.~67, no.~3, pp. 209--221, 2016.

\bibitem{Lourenço2003}
\BIBentryALTinterwordspacing
H.~R. Louren{\c{c}}o, O.~C. Martin, and T.~St{\"u}tzle, \emph{Iterated Local Search}.\hskip 1em plus 0.5em minus 0.4em\relax Boston, MA: Springer US, 2003, pp. 320--353. [Online]. Available: \url{https://doi.org/10.1007/0-306-48056-5_11}
\BIBentrySTDinterwordspacing

\bibitem{Hansen}
P.~Hansen, N.~Mladenovi{\'c}, and J.~A. Moreno~Perez, ``Variable neighbourhood search: methods and applications,'' \emph{Annals of Operations Research}, vol. 175, pp. 367--407, 2010.

\bibitem{ARINGHIERI2016359}
\BIBentryALTinterwordspacing
R.~Aringhieri, A.~Grosso, and P.~Hosteins, ``A genetic algorithm for a class of critical node problems,'' \emph{Electronic Notes in Discrete Mathematics}, vol.~52, pp. 359--366, 2016, iNOC 2015 – 7th International Network Optimization Conference. [Online]. Available: \url{https://www.sciencedirect.com/science/article/pii/S157106531630052X}
\BIBentrySTDinterwordspacing

\bibitem{ARINGHIERI2016128}
``A general evolutionary framework for different classes of critical node problems,'' \emph{Engineering Applications of Artificial Intelligence}, vol.~55, pp. 128--145, 2016.

\bibitem{zhou2018memetic}
Y.~Zhou, J.-K. Hao, and F.~Glover, ``Memetic search for identifying critical nodes in sparse graphs,'' \emph{IEEE transactions on cybernetics}, vol.~49, no.~10, pp. 3699--3712, 2018.

\bibitem{krasnogor2005tutorial}
N.~Krasnogor and J.~Smith, ``A tutorial for competent memetic algorithms: model, taxonomy, and design issues,'' \emph{IEEE transactions on Evolutionary Computation}, vol.~9, no.~5, pp. 474--488, 2005.

\bibitem{zhou2020variable}
Y.~Zhou, J.-K. Hao, Z.-H. Fu, Z.~Wang, and X.~Lai, ``Variable population memetic search: A case study on the critical node problem,'' \emph{IEEE Transactions on Evolutionary Computation}, vol.~25, no.~1, pp. 187--200, 2020.

\bibitem{bavelas1950communication}
A.~Bavelas, ``Communication patterns in task-oriented groups,'' \emph{The journal of the acoustical society of America}, vol.~22, no.~6, pp. 725--730, 1950.

\bibitem{1998Collective}
D.~J. Watts and S.~H. Strogatz, ``Collective dynamics of `small-world' networks,'' \emph{Nature}, vol. 393, pp. 440--442, 1998.

\bibitem{LIN2024202}
F.~Lin, H.~Zhang, J.~Wang, and J.~Wang, ``Unsupervised image enhancement under non-uniform illumination based on paired cnns,'' \emph{Neural Networks}, vol. 170, pp. 202--214, 2024.

\bibitem{kipf2016semi}
T.~N. Kipf and M.~Welling, ``Semi-supervised classification with graph convolutional networks,'' in \emph{Proceedings of the International Conference on Learning Representations(ICLR)}, 2017.

\bibitem{LiuSCIS2023}
C.~Liu, G.~Liu, C.~Luo, S.~Cai, Z.~Lei, W.~Zhang, Y.~Chu, and G.~Zhang, ``Optimizing local search-based partial maxsat solving via initial assignment prediction,'' \emph{SCIENCE CHINA Information Sciences}, pp.~--, 2023.

\bibitem{velickovic2017graph}
P.~Velickovic, G.~Cucurull, A.~Casanova, A.~Romero, P.~Lio, and Y.~Bengio, ``Graph attention networks,'' \emph{stat}, vol. 1050, p.~20, 2017.

\bibitem{vaswani2017attention}
A.~Vaswani, N.~Shazeer, N.~Parmar, J.~Uszkoreit, L.~Jones, A.~N. Gomez, {\L}.~Kaiser, and I.~Polosukhin, ``Attention is all you need,'' \emph{Advances in neural information processing systems}, vol.~30, 2017.

\bibitem{ZHAO202018}
G.~Zhao, P.~Jia, A.~Zhou, and B.~Zhang, ``Infgcn: Identifying influential nodes in complex networks with graph convolutional networks,'' \emph{Neurocomputing}, vol. 414, pp. 18--26, 2020.

\bibitem{clevert2015fast}
D.-A. Clevert, T.~Unterthiner, and S.~Hochreiter, ``Fast and accurate deep network learning by exponential linear units (elus),'' in \emph{Proceedings of the International Conference on Learning Representations(ICLR)}, 2016.

\bibitem{9359655}
M.~Zhang, H.~Li, S.~Pan, J.~Lyu, S.~Ling, and S.~Su, ``Convolutional neural networks-based lung nodule classification: A surrogate-assisted evolutionary algorithm for hyperparameter optimization,'' \emph{IEEE Transactions on Evolutionary Computation}, vol.~25, no.~5, pp. 869--882, 2021.

\bibitem{2014Adam}
D.~P. Kingma and J.~Ba, ``Adam: A method for stochastic optimization,'' in \emph{Proceedings of the 3rd International Conference on Learning Representations}, 2014.

\bibitem{7440882}
Y.~Chen and J.-K. Hao, ``Memetic search for the generalized quadratic multiple knapsack problem,'' \emph{IEEE Transactions on Evolutionary Computation}, vol.~20, no.~6, pp. 908--923, 2016.

\bibitem{aringhieri2016general}
R.~Aringhieri, A.~Grosso, P.~Hosteins, and R.~Scatamacchia, ``A general evolutionary framework for different classes of critical node problems,'' \emph{Engineering Applications of Artificial Intelligence}, vol.~55, pp. 128--145, 2016.

\bibitem{SciPyProceedings_11}
A.~A. Hagberg, D.~A. Schult, and P.~J. Swart, ``Exploring network structure, dynamics, and function using networkx,'' in \emph{Proceedings of the 7th Python in Science Conference}, G.~Varoquaux, T.~Vaught, and J.~Millman, Eds., Pasadena, CA USA, 2008, pp. 11 -- 15.

\bibitem{10.5555/1248547.1248548}
J.~Dem\v{s}ar, ``Statistical comparisons of classifiers over multiple data sets,'' \emph{J. Mach. Learn. Res.}, vol.~7, p. 1–30, dec 2006.

\bibitem{LIU2023}
D.-X. Liu and C.~Qian, ``Result diversification with negative type distances by multi-objective evolutionary algorithms,'' \emph{Fundamental Research}, 2023.

\end{thebibliography}
\end{document}